\documentclass[draftcls]{IEEEtran}

\IEEEoverridecommandlockouts

\usepackage{graphicx}
\usepackage{amsmath,amsfonts}
\usepackage{algorithm}
\usepackage{algorithmicx}
\usepackage{algpseudocode}
\usepackage{balance}
\usepackage{xcolor}
\usepackage[hidelinks]{hyperref}
\usepackage{comment}
\usepackage{orcidlink}

\newif\ifhl
\hltrue
\newcommand{\new}[1]{\ifhl{\color{black}#1}\else#1\fi}

\onecolumn
\title{Robust Covariance Estimation for \new{Entry-wise} Contaminated and Partially Observed Data via Expectation Maximization Algorithm}

\author{
Matys~Savidan$^+$ \orcidlink{0009-0008-7605-2815},
Mohammed~Nabil~El~Korso$^\ddag$ \orcidlink{0000-0002-5489-4308}%
\thanks{$^+$ENS Paris-Saclay, France, $^\ddag$ Paris Saclay University,  France. This work was supported by DGA/AID projects under Grant 2025.65.0026.}
}

\begin{document}
\maketitle

\begin{abstract}
Robust covariance estimation under \new{entry-wise} contamination and missing data is challenging in various signal processing applications. Under the Student-$t$ model, observation-level reweighting induces a uniform attenuation across samples, which can lead to cumulative information loss even when most coordinates remain informative. The alternative Student-$t^*$ model enables coordinate-wise robustness, but \new{its E-step is not available in closed form.}
\new{We extend} inference for the Student-$t^*$ model \new{to} incomplete observations through a stochastic approximation EM (SAEM) algorithm operating on a reduced latent space: stochastic updates of the coordinate-wise scales are combined with Rao--Blackwellized Gaussian conditional moments, which avoids explicit sampling of latent Gaussian variables while preserving a likelihood-based framework.
\new{Experiments on synthetic and real sensor data show a lower Fisher--Rao error than Gaussian, Student-$t$, imputation-based and detect-then-estimate estimators at moderate to high outlier rates under diffuse contamination.} \new{The code is available \href{https://github.com/matyssavidan/Robust-Covariance-Estimation-for-Entry-wise-Contaminated-and-Partially-Observed-Data}{here}.}
\end{abstract}

\section{Introduction}
Likelihood-based covariance estimation in the presence of latent variables or missing data is commonly addressed using the expectation--maximization (EM) algorithm \cite{dempster1977em}\new{, whose efficiency relies on the conditionally Gaussian structure of the augmented model}. However, Gaussian models are inherently
sensitive to model misspecification and data corruption \cite{liu2019parameter,zoubir2012robust}. In such settings, robustness mechanisms that discard entire observations can lead to a significant loss of information.

A widely used robust alternative is based on the multivariate Student-$t$
distribution, which admits a latent scale representation \cite{meriaux2021matched} and can therefore be
handled within an EM framework \cite{liu1995mlt,zhou2020student}. By introducing an observation-level scale variable, this model provides robustness to outliers and heavy-tailed noise \cite{delmas2021elliptical,ollila2012complex}. In the presence of \new{entry-wise contamination, where individual entries are corrupted independently \cite{alqallaf2009propagation,haardt2017subspace}}, the classical Student-$t$ model applies observation-level attenuation, whereby a single corrupted coordinate affects the weight of the entire observation, resulting in the loss of statistical information.

To address \new{such contamination}, Finegold and Drton \cite{finegold2011robust} introduced
the so-called alternative multivariate Student-$t^*$ distribution, which associates an
independent latent scale variable with each coordinate. This formulation enables
localized robustness by selectively down-weighting corrupted entries. Such
\new{contamination} commonly arises in signal processing applications such as
source localization with partially corrupted subarrays or radio interferometry
affected by localized radio-frequency interference
\cite{pesavento2002direction,wijnholds2009multisource}.

However, the increased flexibility of the Student-$t^*$ model comes at the cost of a much more intricate
latent structure. \new{The scales are independent a priori but coupled a posteriori, so that the E-step is not available in closed form, which precludes a standard EM}
\cite{wijnholds2010calibration,wang2025lowrank,gu2012robust}. \new{Likelihood-based inference remains possible through a Monte Carlo approximation of the E-step, but} this
limitation becomes critical in the presence of missing observations, a common
scenario arising from sensor failures or outages, for which no likelihood-based
EM or SAEM inference procedure is currently available for the Student-$t^*$
model \new{\cite{nguyen2021robust,hippertferrer2026missing}}.

\new{A closely related problem is addressed by Lounici and Pacreau \cite{lounici2023robust}, who combine an inverse-probability debiased moment estimator, which requires no imputation, with an external entry-wise detection step. Their approach takes a hard decision per entry, whereas ours never does: corrupted entries are down-weighted continuously, with weights inferred jointly with the parameters, and no detection threshold is needed. Section~\ref{sec:res} compares the two and locates the contamination regime in which each is preferable.}

\new{This letter extends likelihood-based inference for the Student-$t^*$ model to incomplete observations through a SAEM procedure \cite{kuhn2005maximum,mclachlan2008em} in which stochastic approximation is restricted to the coordinate-wise scales, the latent Gaussian components being integrated out analytically. To our knowledge, it is the first likelihood-based EM-type procedure for the Student-$t^*$ model that accommodates missing entries. The Student-$t^*$ model and the Monte Carlo E-step are due to \cite{finegold2011robust}. New here are the extension to incomplete observations through a nested expectation, the Rao--Blackwellization of the Gaussian moments, the stochastic approximation of the scale statistics, the estimation of the degrees of freedom from observed entries only, and an explicit ECM maximization step.}

\section{Alternative Student-$t^*$ Hierarchical Model}

We consider the alternative multivariate Student-$t^*$ distribution introduced by
\cite{finegold2011robust}, which provides coordinate-wise robustness through a
hierarchical formulation.

Let $i = 1,\dots,n$ index independent observations and $j = 1,\dots,p$ denote
the coordinates. The model is defined hierarchically as follows\new{:}
for each observation $i$, we introduce a latent Gaussian random vector
\[
\boldsymbol{x}_i \sim \mathcal{N}_p(\boldsymbol{0},\boldsymbol{\Sigma}),
\]
where $\boldsymbol{\Sigma} \in \mathbb{R}^{p \times p}$ is a positive definite covariance
matrix.
\new{Then,} we associate an independent positive latent scale variable with each
coordinate,
\[
\tau_{ij} \sim \Gamma\!\left(\frac{\nu}{2},\frac{\nu}{2}\right),
\qquad j = 1,\dots,p, \ \  i = 1,\dots,n,
\]
The variables $\tau_{ij}$ are assumed independent.
Finally, the observed data vector $\mathbf{y}_i \in \mathbb{R}^p$ is defined
component-wise by
\begin{equation}
    Y_{ij} = \mu_j + \frac{X_{ij}}{\sqrt{\tau_{ij}}},
\qquad j = 1,\dots,p,
\label{eq:1}
\end{equation}
where $\boldsymbol{\mu} = (\mu_1,\dots,\mu_p)^\top \in \mathbb{R}^p$ is a location parameter. Missing observations may
occur in the components of $\boldsymbol{y}_i$, while the latent variables $\boldsymbol{x}_i$ and the collection $\{\tau_{ij}\}_{i,j}$ remain unobserved.
We denote $\boldsymbol{y}_i \sim t^{\ast}_{p,\nu}(\boldsymbol{\mu},\boldsymbol{\Sigma})$.
\new{The parameter of interest, commonly considered in signal processing applications \cite{ollila2012complex,delmas2021elliptical}, is the covariance matrix $\boldsymbol{\Sigma}$ of the latent Gaussian signal $\boldsymbol{x}_i$, which describes its underlying correlation structure independently of random amplitude variations
\cite{hippertferrer2022robust}. The observed data are affected by an additional random scaling factor, so that their covariance reflects both the signal structure and these amplitude fluctuations. For example, in radar processing, $\boldsymbol{\Sigma}$ characterizes the spatial correlations of the signal received by the different antennas, while random variations in target or clutter power change the observed covariance. We therefore estimate the underlying covariance matrix rather than the covariance of the scaled observations.}

\section{Robust Inference with Incomplete Observations}

\subsection{Latent structure and complete-data likelihood}

For each observation $i$, let $\mathcal O_i$ and $\mathcal M_i$ denote the sets of observed and missing coordinates. The available data consist of $\boldsymbol{Y}_{\mathcal O} = \{Y_{\mathcal O_i}\}_{i=1}^n$.
Conditional on the scales $\boldsymbol{\tau}$, the Gaussian vector $\boldsymbol{x}_i$ is fully determined on the observed subset $\mathcal O_i$ (cf.~\eqref{eq:1}), but remains latent on $\mathcal M_i$.
Consequently, the complete data involve the observed entries, the missing values $\boldsymbol{Y}_{\mathcal{M}}$, and the scale variables $\boldsymbol{\tau}$. Assuming an ignorable missingness mechanism, the complete-data log-likelihood
$\ell_c(\boldsymbol{\theta})$, with $\boldsymbol{\theta} = (\boldsymbol{\mu},\boldsymbol{\Sigma},\nu)$, reads
\begin{align}
\ell_c(\boldsymbol{\theta}) = &  \log p(\boldsymbol{Y}_{\mathcal O}, \boldsymbol{Y}_{\mathcal{M}}, \boldsymbol{\tau}  \boldsymbol{\theta}) \nonumber \\ = & \sum_{i=1}^n \Bigg[  -\frac{1}{2}\ln|\boldsymbol{\Sigma}|
   - \frac{1}{2}
 (D(\boldsymbol{\tau}_i)^{1/2}(\boldsymbol{y}_i - \boldsymbol{\mu}))^\top\boldsymbol{\Sigma}^{-1}
  (D(\boldsymbol{\tau}_i)^{1/2}(\boldsymbol{y}_i - \boldsymbol{\mu}))
\nonumber \\
&  + \sum_{j=1}^p \Big( \frac{\nu}{2}\ln\frac{\nu}{2} - \ln\Gamma\!\Big(\frac{\nu}{2}\Big)     + \Big(\frac{\nu}{2}-1\Big)\ln\tau_{ij} - \frac{\nu}{2}\tau_{ij} \Big)  
 + \frac{1}{2} \sum_{j=1}^p \ln \tau_{ij} \Bigg] + \text{const},
\label{eq:complete_ll}
\end{align}
in which $D(\boldsymbol{\tau}_i) = \mathrm{diag}(\tau_{i1}, \dots, \tau_{ip})$, the latent vector can be rewritten as
$\boldsymbol{x}_i = D(\boldsymbol{\tau}_i)^{1/2}(\boldsymbol{y}_i - \boldsymbol{\mu})$\new{, and the constant does not depend on $\boldsymbol{\theta}$}.

\subsection{EM formulation and sufficient statistics}

Inference relies on maximizing the observed-data log-likelihood $\ell(\boldsymbol{\theta}) = \log p(\boldsymbol{Y}_O \mid \boldsymbol{\theta})$. This is addressed via the EM algorithm by iteratively maximizing the surrogate function $Q(\boldsymbol{\theta}\mid\boldsymbol{\theta}^{(k)})$. Exploiting the hierarchical structure, the expectation step decomposes as:
\begin{align}
Q(\boldsymbol{\theta}\mid\boldsymbol{\theta}^{(k)})
&=
\mathbb{E}_{\boldsymbol{\tau},\boldsymbol{Y}_{\mathcal M} \mid \boldsymbol{Y}_O,\boldsymbol{\theta}^{(k)}}
\!\left[\ell_c(\boldsymbol{\theta})\right] \nonumber\\
&=
\mathbb{E}_{\boldsymbol{\tau} \mid \boldsymbol{Y}_O,\boldsymbol{\theta}^{(k)}}
\!\left[
\mathbb{E}_{\boldsymbol{Y}_{\mathcal M}\mid \boldsymbol{Y}_O,\boldsymbol{\tau},\boldsymbol{\theta}^{(k)}}
\!\left[\ell_c(\boldsymbol{\theta})\right]
\right].
\label{eq:Q_decomp}
\end{align}
The inner expectation over missing data is analytically tractable conditional on $\boldsymbol{\tau}$; the outer one is over the scales. Let $\mathbb{E}[\cdot]$ denote expectation with respect to $p(\boldsymbol{\tau}, \boldsymbol{Y}_{\mathcal M} \mid \boldsymbol{Y}_{\mathcal O}, \theta^{(k)})$. Maximizing \eqref{eq:Q_decomp} relies on the conditional expectations of the sufficient statistics:
\begin{align}
\bar{S}_{\Sigma}
&= \frac{1}{n}\sum_{i=1}^n
\mathbb{E}\!\left[ \boldsymbol{x}_i \boldsymbol{x}_i^\top \right],
\label{eq:suff_sigma} \\
\bar{S}_{\mu,1}
&= \sum_{i=1}^n
\mathbb{E}\!\left[ W_i \,y_i \right], \quad
\bar{S}_{\mu,2}
= \sum_{i=1}^n
\mathbb{E}\!\left[ W_i \right],
\label{eq:suff_mu} \\
\bar{S}_{\tau}
&= \sum_{i=1}^n \sum_{j=1}^p
\mathbb{E}\!\left[ \tau_{ij} \right], \quad
\bar{S}_{\log \tau}
= \sum_{i=1}^n \sum_{j=1}^p
\mathbb{E}\!\left[ \log \tau_{ij} \right],
\label{eq:suff_nu}
\end{align}
where $W_i = D(\boldsymbol{\tau}_i)^{1/2}\, \boldsymbol{\Sigma}^{-1}\, D(\boldsymbol{\tau}_i)^{1/2}$.

Evaluating these expectations is non-trivial. The inner expectation over $\boldsymbol{Y}_{\mathcal M}$ admits a closed form via Gaussian conditioning identities (since $\boldsymbol{x}_i$ is Gaussian given $\boldsymbol{\tau}_i$), \new{but the posterior of $\boldsymbol{\tau}_i$ does not \cite{finegold2011robust}: the $\tau_{ij}$ are independent a priori but coupled a posteriori through non diagonal $\boldsymbol{\Sigma}_{\mathcal O_i\mathcal O_i}^{-1}$, which rules out a factorization, and the integral is $p$-dimensional, so that numerical quadrature is impractical. The E-step therefore has no closed form. Direct maximization of $\ell$ does not avoid it: by Fisher's identity, $\nabla\ell(\boldsymbol{\theta})=\mathbb{E}[\nabla\ell_c(\boldsymbol{\theta})\mid\boldsymbol{Y}_O]$, so a gradient method needs the same posterior expectation.} This motivates the use of the Stochastic Approximation EM (SAEM) described in Section IV.

\section{Stochastic Approximation EM Algorithm}

\subsection{Rao--Blackwellized Gaussian Moments}
\label{sec:RB}
Rather than sampling $\mathbf{x}_i$, we compute its conditional moments analytically, which gives a Rao--Blackwellized implementation: conditionally on $\boldsymbol{\tau}_i$ and $Y_{i,O_i}$,
\begin{equation}
\mathbf{x}_i \mid Y_{i,O_i}, \boldsymbol{\tau}_i
\sim \mathcal{N}(\mathbf{m}_i,\mathbf{C}_i),
\label{eq:x_conditional}
\end{equation}
with
\begin{equation}
\mathbf{m}_i = \boldsymbol{\Sigma}_{:,O_i} \boldsymbol{\Sigma}_{O_iO_i}^{-1} \mathbf{D}(\boldsymbol{\tau}_{i,O_i})^{1/2} (\mathbf{y}_{i,O_i}-\boldsymbol{\mu}_{O_i})
\label{eq:conditional_mean}
\end{equation}
\begin{equation}
\mathbf{C}_i = \boldsymbol{\Sigma} - \boldsymbol{\Sigma}_{:,O_i} \boldsymbol{\Sigma}_{O_iO_i}^{-1} \boldsymbol{\Sigma}_{O_i, :}.
\label{eq:conditional_cov}
\end{equation}
\new{When all coordinates are observed, $\mathbf{C}_i=\boldsymbol{0}$ and $\mathbf{m}_i=\mathbf{x}_i$: the Rao--Blackwellization is then void and the method reduces to a SAEM for the same model.}
These expressions give exactly all Gaussian-related conditional
moments required for the sufficient statistics of $(\boldsymbol{\mu},\boldsymbol{\Sigma})$:
\begin{small}
\begin{align}
\bar{S}_\Sigma
& = \frac{1}{n}\sum_{i=1}^n \mathbb{E}_{\tau}\!\left[ \boldsymbol{C}_i + \boldsymbol{m}_i \boldsymbol{m}_i^\top \right], \label{eq:SbarSigma}\\
\bar{S}_{\mu,1}
&= \sum_{i=1}^n \mathbb{E}_{\tau}\!\left[
W_i \big( \boldsymbol{\mu} + D(\boldsymbol{\tau}_i)^{-1/2} \boldsymbol{m}_i \big)
\right].
\label{eq:esp}
\end{align}
\end{small}

\subsection{Sampling of the scale variables}

At each iteration, for observed coordinates ($j \in O_i$), $\tau_{ij}$ is updated coordinate-wise by a Metropolis--Hastings step targeting its full conditional distribution given $\boldsymbol{\tau_{i,-j}}$ and the observed data $\boldsymbol{y_{\mathcal O_i}}$,
\[
\pi(\tau_{ij} \mid \boldsymbol{\tau_{i,-j}}, \boldsymbol{y_{\mathcal O_i}})
\propto p(\tau_{ij}\mid \nu)\; p(\boldsymbol{y_{\mathcal O_i}} \mid \boldsymbol{\tau_i}).
\]
\new{Since $\boldsymbol{y}_{\mathcal O_i}\mid\boldsymbol{\tau}_i$ is Gaussian with mean $\boldsymbol{\mu}_{\mathcal O_i}$ and covariance $D^{-1/2}\boldsymbol{\Sigma}_{\mathcal O_i\mathcal O_i}D^{-1/2}$, $D=D(\boldsymbol{\tau}_{i,\mathcal O_i})$,}
\begin{align}
  \log p(\boldsymbol{y}_{\mathcal O_i}\mid\boldsymbol{\tau}_i) =   -\tfrac12\log|\boldsymbol{\Sigma}_{\mathcal O_i\mathcal O_i}| + \tfrac12\!\sum_{j\in\mathcal O_i}\!\log\tau_{ij}  - \tfrac12\, \boldsymbol{v}_i^\top\boldsymbol{\Sigma}_{\mathcal O_i\mathcal O_i}^{-1}\boldsymbol{v}_i + \text{const},
\label{eq:loglik_y}
\end{align}
\new{with $\boldsymbol{v}_i=D(\boldsymbol{\tau}_{i,\mathcal O_i})^{1/2}(\boldsymbol{y}_{i,\mathcal O_i}-\boldsymbol{\mu}_{\mathcal O_i})$.}
A log-normal random walk proposal is used, $\tau'_{ij} = \tau_{ij} \exp(\eta)$, $\eta \sim \mathcal{N}(0, s_p^2)$, with acceptance probability
$\alpha = \min\big(1,\,\pi(\tau'_{ij})\tau'_{ij}/(\pi(\tau_{ij})\tau_{ij})\big)$.
\new{The proposal scale $s_p$ is adapted by a Robbins--Monro recursion on $\log s_p$ during burn-in, targeting a $40\%$ acceptance rate (about $44\%$ in our experiments), and is then frozen so that the chain keeps the correct invariant distribution.}
For missing coordinates ($j \in M_i$), the likelihood does not depend on
$\tau_{ij}$ and the update reduces to sampling from the Gamma prior.

\subsection{Stochastic Approximation and Parameter Updates}

At iteration $k$, $\mathbb{E}_{\tau}[\cdot]$ is replaced by the Monte Carlo
sample $\tau^{(k)}$, yielding empirical statistics
$\hat{\mathbf{S}}^{(k)}$, and the sufficient statistics are updated by
\begin{equation}
\mathbf{S}^{(k)} = (1-\gamma_k)\mathbf{S}^{(k-1)} + \gamma_k \hat{\mathbf{S}}^{(k)},
\label{eq:sa_update}
\end{equation}
where $(\gamma_k)$ is a decreasing sequence of step sizes.

\subsubsection{\new{Maximization step as an ECM procedure}}

\new{Since $\boldsymbol{\mu}$ and $\boldsymbol{\Sigma}$ are not decoupled ($\boldsymbol{x}_i$ depends on $\boldsymbol{\mu}$ and $W_i$ on $\boldsymbol{\Sigma}$), we adopt a block coordinate ascent approach, namely an Expectation Conditional Maximization (ECM) step \cite{meng1993ecm}: each conditional maximizer is exact when the other parameter is held fixed.}

\new{\emph{CM1 (location).} With $\boldsymbol{\Sigma}^{(k)}$ fixed, $\boldsymbol{\mu}$ is updated by weighted least squares,}
\begin{equation}
 \boldsymbol{\mu}^{(k+1)}
= \left(\mathbf{S}_{\mu,2}^{(k)}\right)^{-1} \mathbf{S}_{\mu,1}^{(k)},
\label{eq:majmu}
\end{equation}
\new{where $\mathbf{S}_{\mu,1}^{(k)}$, $\mathbf{S}_{\mu,2}^{(k)}$ follow \eqref{eq:sa_update} from the sampled $\boldsymbol{\tau}^{(k)}$ using \eqref{eq:suff_mu} and \eqref{eq:esp}.}

\new{\emph{CM2 (covariance).} The moments $\mathbf{m}_i^{(k)},\mathbf{C}_i^{(k)}$ are recomputed at $\boldsymbol{\mu}^{(k+1)}$ (at negligible cost, $\boldsymbol{\Sigma}_{\mathcal O_i\mathcal O_i}^{-1}$ being unchanged), giving the Monte Carlo estimate}
\begin{equation}
  \hat{\mathbf{S}}_{\Sigma}^{(k)}
= \frac{1}{n}\sum_{i=1}^n
\left( \mathbf{C}_i^{(k)} + \mathbf{m}_i^{(k)} {\mathbf{m}_i^{(k)}}^\top \right),
\label{eq:sigma_innov}
\end{equation}
\new{and, after \eqref{eq:sa_update}, $\boldsymbol{\Sigma}^{(k+1)} = \mathbf{S}_{\Sigma}^{(k)}$. Each step is an exact conditional maximizer, so the maximization is a generalized EM step.}

\subsubsection{Degrees of Freedom}

The parameter $\nu$ is updated by solving
\begin{equation}
\log(\nu/2) - \psi(\nu/2) + 1
+ \frac{1}{N}\big(S_{\ln \tau}^{(k)} - S_{\tau}^{(k)}\big) = 0,
\label{eq:nu_update}
\end{equation}
where $N = \sum_i |O_i|$. $S_{\tau}^{(k)}$ and
$S_{\ln \tau}^{(k)}$ follow \eqref{eq:sa_update} and
use observed entries only, as scales of missing
coordinates are sampled from the prior and carry no information on $\nu$.

\vspace{2pt}\noindent\emph{Algorithm.} Parameters are initialized coordinate-wise using classical estimators: median and nanvar. \new{As $\boldsymbol{\mu},\boldsymbol{\Sigma}$ are frozen during the simulation step, $\boldsymbol{\Sigma}_{\mathcal O_i\mathcal O_i}^{-1}$ is computed once per observation and a single-entry proposal updates $\boldsymbol{v}_i^\top\boldsymbol{\Sigma}_{\mathcal O_i\mathcal O_i}^{-1}\boldsymbol{v}_i$ in $O(1)$: the cost per iteration is $O(np^3+Knp^2)$ for $K$ sweeps.}

\begin{algorithm}
\footnotesize
\begin{algorithmic}[1]
\State Initialize $\boldsymbol{\mu}^{(0)},\boldsymbol{\Sigma}^{(0)},\boldsymbol{\tau}^{(0)},\nu^{(0)}$
\For{$k=1, \dots, K_{max}$}
  \State $\gamma_k \leftarrow (k - K_{burn})^{-\alpha}$ if $k > K_{burn}$ else $1$
  \State \textbf{Simulation:} draw $\tau_{ij}^{(k)} \sim p(\tau_{ij} \mid y_{\mathcal{O}_i}, \theta^{(k-1)})$ via MH, $\forall i, j \in \mathcal{O}_i$
  \State \textbf{Rao--Blackwellization:} compute $\mathbf{m}_i^{(k)}, \mathbf{C}_i^{(k)}$ via \eqref{eq:conditional_mean}--\eqref{eq:conditional_cov}, $\forall i$
  \State \new{\textbf{SA and CM1:} update $\mathbf{S}_{\mu}^{(k)}, \mathbf{S}_{\nu}^{(k)}$ via \eqref{eq:sa_update}; $\boldsymbol{\mu}^{(k+1)} \leftarrow$ Solve \eqref{eq:majmu}}
  \State \new{\textbf{CM2:} recompute $\mathbf{m}_i^{(k)}, \mathbf{C}_i^{(k)}$ at $\boldsymbol{\mu}^{(k+1)}$; update $\mathbf{S}_{\Sigma}^{(k)}$ via \eqref{eq:sa_update}, \eqref{eq:sigma_innov}; $\boldsymbol{\Sigma}^{(k+1)} \leftarrow \mathbf{S}_{\Sigma}^{(k)}$}
  \State \new{$\nu^{(k+1)} \leftarrow$ Solve \eqref{eq:nu_update}}
\EndFor
\end{algorithmic}
\end{algorithm}

\section{Results}
\label{sec:res}
 
\begin{figure}
    \centering
     \includegraphics[width=\textwidth]{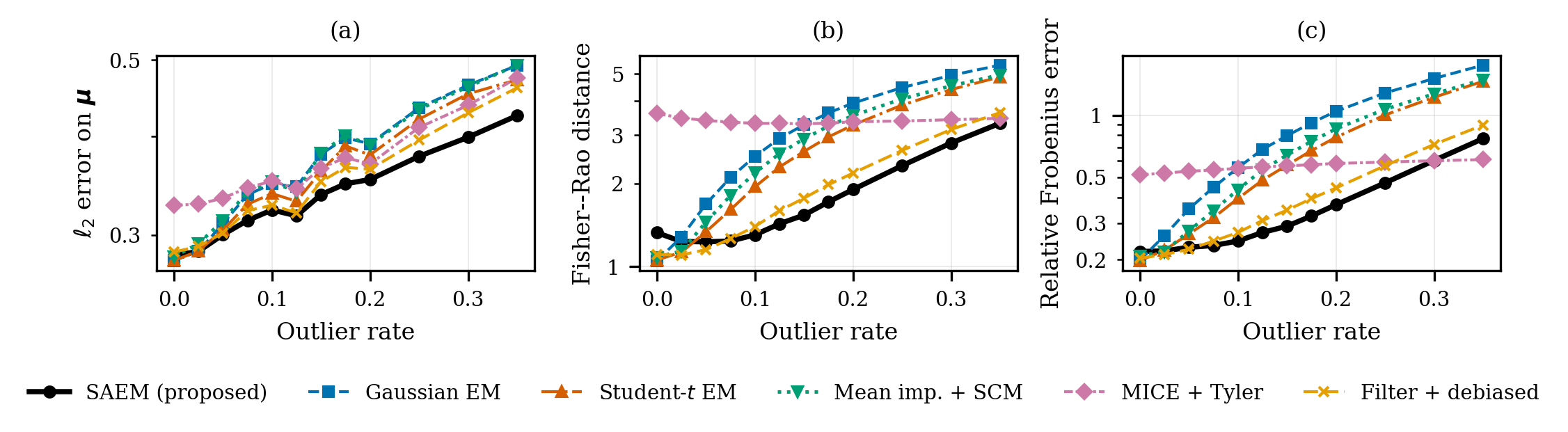}
    \caption{{(a) $\ell_2$ estimation error of the mean vector $\boldsymbol{\mu}$, (b) Fisher--Rao distance and (c) relative Frobenius error of the estimated covariance matrix $\boldsymbol{\Sigma}$ versus outlier rate ($n=500$, $p=20$, $s=2.5$, 10\% missing entries, 20 replications). Compared methods: SAEM (proposed), Gaussian EM~\cite{dempster1977em}, Student-$t$ EM~\cite{liu1995mlt}, mean imputation + SCM~\cite{little2019statistical}, MICE + Tyler~\cite{vanbuuren2011mice,tyler1987distribution}, filter + debiased~\cite{lounici2023robust}.}}
    \label{fig:fisher}
\end{figure}

\subsection{Synthetic data}
Synthetic data are generated from a multivariate Gaussian distribution with
sample size $n=500$ and dimension $p=20$. \new{Each replication draws its own ground-truth covariance $\boldsymbol{\Sigma}$, shared by all outlier rates and estimators (common random numbers), so that comparisons are paired.} Missing entries are introduced uniformly at random, and contamination \new{follows the independent contamination model \cite{alqallaf2009propagation}: each entry is independently replaced, with probability equal to the outlier rate,} by a draw from $\mathcal{N}(\mu_j, s^2 \sigma_j^2)$, where $\sigma_j$ is the marginal standard deviation of coordinate $j$ and $s$ a scale factor, resulting in localized outliers. \new{We take $s=2.5$ (diffuse contamination, outliers within the bulk) and average over 20 replications.}
 
The proposed SAEM is compared with a Gaussian EM estimator\new{~\cite{dempster1977em,little2019statistical}}, the classical multivariate
Student-$t$ EM\new{~\cite{liu1995mlt,little1988robust}} based on observation-level down-weighting, and covariance
estimators built from mean or MICE imputation\new{~\cite{little2019statistical,vanbuuren2011mice}} followed by \new{the sample covariance or Tyler's estimator~\cite{tyler1987distribution} (trace-normalized, hence not estimating the scale). A detect-then-estimate pipeline in the spirit of \cite{lounici2023robust} is also considered: a univariate filter (median and MAD~\cite{rousseeuw1993alternatives}, $3\sigma$) turns flagged entries into missing values, followed by a debiased moment estimator~\cite{lounici2014high,lounici2023robust}.} Performance is assessed using the relative $\ell_2$ error
for the mean estimate and the Frobenius and Fisher--Rao distances for the
covariance matrix, the latter capturing geometric distortions beyond entrywise
discrepancies\new{: $d_{\mathrm{FR}}^2=\sum_i\log^2\lambda_i$, with $\lambda_i$ the eigenvalues of $\boldsymbol{\Sigma}^{-1}\hat{\boldsymbol{\Sigma}}$. It splits into a scale term $p(\log c^\star)^2$, where $c^\star=(\det\hat{\boldsymbol{\Sigma}}/\det\boldsymbol{\Sigma})^{1/p}$ is the overall scale factor of the estimate ($c^\star=1$ for a correctly scaled estimate), and a shape term}.
 
\new{Above about $8\%$ outliers, SAEM has the lowest Fisher--Rao distance of all estimators, detection included, and its gap with the Gaussian and Student-$t$ EM widens with the outlier rate (Fig.~\ref{fig:fisher}(b)). The relative Frobenius error shows the same ranking from about $7\%$ up to about $30\%$ outliers, beyond which the trace-normalized MICE + Tyler, whose error is almost constant, becomes better (Fig.~\ref{fig:fisher}(c)), and the error on the mean is the lowest from about $10\%$ outliers (Fig.~\ref{fig:fisher}(a)). A similar ranking is observed with $20\%$ missing entries. At lower outlier rates, the detector has a lower Fisher--Rao distance (below about $7\%$), as have the Gaussian and Student-$t$ EM without contamination: SAEM pays the usual price of robustness.}
 
\new{The Fisher--Rao distance of SAEM is not monotone: the slight decrease at low outlier rates is due to a scale-calibration bias that changes sign (the scale factor $c^\star$ equals one at about $9\%$ outliers), whereas the shape component increases with the outlier rate. The estimated $\hat\nu$ falls from $13.8$ without contamination to $5$--$7$ for $10$--$35\%$ outliers, against $13$--$18$ for the Student-$t$ EM, which cannot attribute the corruption to single entries: the tail parameter is an interpretable indicator of contamination. The likelihood formulation also requires no detection threshold, provides the conditional distribution of the missing entries as a by-product, and handles missing data jointly with the contamination: applying the same algorithm to mean- or MICE-imputed data instead of the incomplete data increases the Fisher--Rao distance at every missing rate (at $20\%$ missing entries: $2.43$ and $2.67$ against $1.44$), since single imputation discards the conditional covariance $\mathbf{C}_i$.}
 
\new{The contamination magnitude $s$ determines whether detection can solve the problem. For large, separable outliers ($s=10$), the filter flags most corrupted entries and the detect-then-estimate pipeline outperforms all likelihood-based estimators; the regime in which a threshold-free approach is relevant is diffuse contamination: at $s=2.5$ the filter flags $1.9\%$ of the entries whereas $9\%$ are corrupted, and SAEM has the lower distance. With $10\%$ outliers and $10\%$ missing entries, the sign of the paired difference between SAEM and the pipeline changes at $s\approx3.6$, consistently with the dependence of \cite{lounici2023robust} on the success of the filtering step. At constant $n/[p(p+1)/2]=2.38$, SAEM keeps the lowest distance for $p$ from $20$ to $60$, at a runtime comparable to that of the Student-$t$ EM for $p=20$ and about five times larger for $p=60$.}
 
\subsection{Real Data}
 
We further evaluate the proposed approach on real temperature measurements from
the Intel Berkeley Research Lab dataset \cite{intel_lab_data}. A subset of \new{five} sensors
with the most complete recordings is selected \new{($n=1000$, condition number $60$ of the clean covariance, a criterion fixed beforehand)} and arranged into a multivariate
time series, with additional \new{entry-wise} contamination \new{($s=2.5$)} and missing entries (\new{10\%})
introduced following the same protocol as in the synthetic experiments \new{(semi-synthetic; reference: empirical covariance of the clean data; 10 draws)}.
 
\begin{table}[t]
\centering
\footnotesize
\setlength{\tabcolsep}{3pt}
\caption{\new{Fisher--Rao distance on real temperature data ($n=1000$, $p=5$, $s=2.5$, 10\% missing).}}
\new{\begin{tabular}{l|cccc}
\hline
Outlier rate   & 5\% & 10\% & 15\% \\
\hline
Gaussian EM   & 2.243 & 3.142 & 3.776\\
Student-$t$ EM   & 4.363 & 2.573 & \textbf{1.010}\\
Filter + debiased   & 1.638 & 2.373 & 2.905\\
MICE + Tyler   & 2.535 & 1.988 & 2.187\\
Mean imp. + SCM   & 2.313 & 3.093 & 3.672\\
SAEM (proposed)   & \textbf{0.792} & \textbf{0.824} & \underline{1.528}\\
\hline
\end{tabular}}
\end{table}
 
\new{Table~I shows that SAEM has the lowest distance at $5\%$ and $10\%$ outliers (about half of the next best), whereas the Student-$t$ EM is the best at $15\%$, and the Gaussian EM without contamination, as expected since the reference is the clean empirical covariance. The advantage of the Student-$t$ EM at $15\%$ is plausibly due to the small dimension: with $p=5$ an observation remains fully uncorrupted with probability $0.85^5\approx44\%$ (against $4\%$ for $p=20$), so that observation-level down-weighting discards little clean information, whereas the benefit of entry-wise scales grows with $p$. The relative Frobenius error and the error on the mean do not favour SAEM on these data. The same ranking is obtained with $20\%$ missing entries; this and further experiments are available in the repository.}
 
\section{Conclusion}
 
This letter introduced a likelihood-based SAEM framework for robust covariance
estimation under \new{entry-wise} contamination and missing data, combining
coordinate-wise scale modeling with Rao--Blackwellized Gaussian updates \new{and an explicit ECM maximization step}. Experiments on synthetic and real data show
\new{that coordinate-wise scales give the most accurate estimates in Fisher--Rao distance when contamination is diffuse and not negligible, whereas a plain detection step remains preferable for very low contamination or for large, separable outliers}. These findings highlight the
benefit of coordinate-level robustness in modern sensing and
signal processing applications.
 
\bibliographystyle{IEEEtran}
\bibliography{references}

@article{finegold2011robust,
  author  = {Finegold, Michael and Drton, Mathias},
  title   = {Robust graphical modeling of gene networks using classical and alternative t-distributions},
  journal = {The Annals of Applied Statistics},
  volume  = {5},
  number  = {2A},
  pages   = {1057--1080},
  year    = {2011},
  doi     = {10.1214/10-AOAS410}
}

@article{liu1995mlt,
  author  = {Liu, Chuanhai and Rubin, Donald B.},
  title   = {ML Estimation of the $t$ Distribution Using EM and Its Extensions, ECM and ECME},
  journal = {Statistica Sinica},
  volume  = {5},
  number  = {1},
  pages   = {19--39},
  year    = {1995}
}

@article{dempster1977em,
  title   = {Maximum Likelihood from Incomplete Data via the EM Algorithm},
  author  = {Dempster, Arthur P. and Laird, Nan M. and Rubin, Donald B.},
  journal = {Journal of the Royal Statistical Society, Series B},
  volume  = {39},
  number  = {1},
  pages   = {1--38},
  year    = {1977}
}

@misc{intel_lab_data,
  title = {Intel Lab Data},
  author = {{Intel Berkeley Research Lab}},
  year = {2004},
  howpublished = {\url{http://db.csail.mit.edu/labdata/labdata.html}},
  note = {Accessed: 2024}
}

@article{zhou2020student,
  author  = {Zhou, R. and Liu, J. and Kumar, S. and Palomar, D.},
  title   = {Student's t VAR modeling with missing data via stochastic EM and Gibbs sampling},
  journal = {IEEE Transactions on Signal Processing},
  volume  = {68},
  pages   = {6198--6211},
  year    = {2020}
}

@book{delmas2021elliptical,
  author    = {Delmas, Jean-Pierre and El Korso, Mohammed Nabil and Pascal, Fr{\'e}d{\'e}ric and Fortunati, Stefano},
  title     = {Elliptically Symmetric Distributions in Signal Processing and Machine Learning},
  publisher = {Springer Nature},
  year      = {2024},
  month     = dec
}

@article{ollila2012complex,
  author  = {Ollila, E. and Tyler, D. E. and Koivunen, V. and Poor, H. V.},
  title   = {Complex elliptically symmetric distributions: Survey, new results and applications},
  journal = {IEEE Transactions on Signal Processing},
  volume  = {60},
  number  = {11},
  pages   = {5597--5625},
  year    = {2012},
  month   = nov
}

@article{pesavento2002direction,
  author  = {Pesavento, M. and Gershman, A. B. and Wong, K. M.},
  title   = {Direction finding in partly calibrated sensor arrays composed of multiple subarrays},
  journal = {IEEE Transactions on Signal Processing},
  volume  = {50},
  number  = {9},
  pages   = {2103--2115},
  year    = {2002}
}

@article{wijnholds2009multisource,
  author  = {Wijnholds, S. J. and Van der Veen, A. J.},
  title   = {Multisource self-calibration for sensor arrays},
  journal = {IEEE Transactions on Signal Processing},
  year    = {2009}
}

@article{wijnholds2010calibration,
  author  = {Wijnholds, S. J. and van der Tol, S. and Nijboer, R. and van der Veen, A. J.},
  title   = {Calibration challenges for future radio telescopes},
  journal = {IEEE Signal Processing Magazine},
  volume  = {27},
  number  = {1},
  pages   = {30--42},
  year    = {2010}
}

@article{wang2025lowrank,
  author  = {Wang, J. and El Korso, M. N. and Bacharach, L. and Larzabal, P.},
  title   = {Low-rank EM-based imaging for large-scale switched interferometric arrays},
  journal = {IEEE Signal Processing Letters},
  volume  = {32},
  pages   = {41--45},
  year    = {2025}
}

@article{gu2012robust,
  author  = {Gu, Y. and Leshem, A.},
  title   = {Robust adaptive beamforming based on interference covariance matrix reconstruction and steering vector estimation},
  journal = {IEEE Transactions on Signal Processing},
  volume  = {60},
  number  = {7},
  pages   = {3881--3885},
  year    = {2012}
}

@article{nguyen2021robust,
  author  = {Nguyen, Viet Dung and Nguyen, Linh Trung and Abed-Meraim, Karim and others},
  title   = {Robust subspace tracking with missing data and outliers: Novel algorithm with convergence guarantee},
  journal = {IEEE Transactions on Signal Processing},
  volume  = {69},
  pages   = {2070--2085},
  year    = {2021}
}

@article{kuhn2005maximum,
  author  = {Kuhn, E. and Lavielle, M.},
  title   = {Maximum likelihood estimation in nonlinear mixed effects models},
  journal = {Computational Statistics \& Data Analysis},
  volume  = {49},
  number  = {4},
  pages   = {1020--1038},
  year    = {2005}
}

@book{mclachlan2008em,
  author    = {McLachlan, G. J. and Krishnan, T.},
  title     = {The {EM} Algorithm and Extensions},
  publisher = {Wiley},
  year      = {2008},
  edition   = {2}
}

@article{liu2019parameter,
  author  = {Liu, J. and Kumar, S. and Palomar, D. P.},
  title   = {Parameter estimation of heavy-tailed {AR} model with missing data via stochastic {EM}},
  journal = {IEEE Transactions on Signal Processing},
  volume  = {67},
  number  = {8},
  pages   = {2159--2172},
  year    = {2019}
}

@article{zoubir2012robust,
  author  = {Zoubir, A. M. and Koivunen, V. and Chakhchoukh, Y. and Muma, M.},
  title   = {Robust estimation in signal processing: A tutorial-style treatment of fundamental concepts},
  journal = {IEEE Signal Processing Magazine},
  volume  = {29},
  number  = {4},
  pages   = {61--80},
  year    = {2012},
  month   = jul
}

@article{HippertFerrer2022Robust,
  title   = {Robust low-rank covariance matrix estimation with a general pattern of missing values},
  author  = {Hippert-Ferrer, A. and El Korso, M. N. and Breloy, A. and Ginolhac, G.},
  journal = {Signal Processing},
  volume  = {195},
  pages   = {108460},
  year    = {2022},
  month   = {June},
  doi     = {10.1016/j.sigpro.2022.108460}
}

@article{alqallaf2009propagation,
  author  = {Alqallaf, Fatemah and Van Aelst, Stefan and Yohai, V{\'\i}ctor J. and Zamar, Ruben H.},
  title   = {Propagation of outliers in multivariate data},
  journal = {The Annals of Statistics},
  volume  = {37}, number = {1}, pages = {311--331}, year = {2009}
}

@inproceedings{lounici2023robust,
  author    = {Lounici, Karim and Pacreau, Gr{\'e}goire},
  title     = {Robust covariance estimation with missing values and cell-wise contamination},
  booktitle = {Advances in Neural Information Processing Systems},
  volume    = {36}, pages = {72124--72136}, year = {2023}
}

@article{meng1993ecm,
  author  = {Meng, Xiao-Li and Rubin, Donald B.},
  title   = {Maximum likelihood estimation via the {ECM} algorithm: A general framework},
  journal = {Biometrika},
  volume  = {80}, number = {2}, pages = {267--278}, year = {1993}
}

@article{hippertferrer2026missing,
  author  = {A. Hippert-Ferrer and A. Sportisse and A. Javaheri and M. N. {El Korso} and D. P. Palomar},
  title   = {Missing Data in Signal Processing and Machine Learning: Models, Methods, and Modern Approaches},
  journal = {IEEE Signal Processing Magazine},
  volume  = {43},
  number  = {4},
  year    = {2026}
}

@book{little2019statistical,
  author    = {Little, Roderick J. A. and Rubin, Donald B.},
  title     = {Statistical Analysis with Missing Data},
  edition   = {3rd},
  publisher = {Wiley},
  year      = {2019}
}

@article{little1988robust,
  author  = {Little, Roderick J. A.},
  title   = {Robust estimation of the mean and covariance matrix from data with missing values},
  journal = {Journal of the Royal Statistical Society: Series C (Applied Statistics)},
  volume  = {37}, number = {1}, pages = {23--38}, year = {1988}
}

@article{vanbuuren2011mice,
  author  = {van Buuren, Stef and Groothuis-Oudshoorn, Karin},
  title   = {{mice}: Multivariate imputation by chained equations in {R}},
  journal = {Journal of Statistical Software},
  volume  = {45}, number = {3}, pages = {1--67}, year = {2011}
}

@article{tyler1987distribution,
  author  = {Tyler, David E.},
  title   = {A distribution-free {M}-estimator of multivariate scatter},
  journal = {The Annals of Statistics},
  volume  = {15}, number = {1}, pages = {234--251}, year = {1987}
}

@article{rousseeuw1993alternatives,
  author  = {Rousseeuw, Peter J. and Croux, Christophe},
  title   = {Alternatives to the median absolute deviation},
  journal = {Journal of the American Statistical Association},
  volume  = {88}, number = {424}, pages = {1273--1283}, year = {1993}
}

@article{lounici2014high,
  author  = {Lounici, Karim},
  title   = {High-dimensional covariance matrix estimation with missing observations},
  journal = {Bernoulli},
  volume  = {20}, number = {3}, year = {2014}
}

@incollection{haardt2017subspace,
  title     = {Subspace methods and exploitation of special array structures},
  author    = {Haardt, M. and Pesavento, M. and Roemer, F. and El Korso, M. N.},
  booktitle = {Academic Press Library in Signal Processing},
  volume    = {3},
  pages     = {651--717},
  publisher = {Academic Press},
  year    = {2017}
}

@article{meriaux2021matched,
  title   = {Matched and mismatched estimation of Kronecker product of linearly structured scatter matrices under elliptical distributions},
  author  = {Meriaux, B. and Ren, C. and Breloy, A. and El Korso, M. N. and Forster, P.},
  journal = {IEEE Transactions on Signal Processing},
  volume  = {69},
  pages   = {603--616},
  year    = {2021}
}
 
\end{document}